\documentclass[runningheads]{llncs}
\usepackage[T1]{fontenc}
\usepackage{graphicx} 
\usepackage{hyperref}       
\usepackage{url}            

\usepackage{amsfonts}       
\usepackage{bbm}
\usepackage{xcolor}         
\usepackage{amsmath}
\usepackage{appendix}
\usepackage{ulem}
\usepackage{multirow}
\usepackage{booktabs}
\usepackage{array,float}
\usepackage{threeparttable}
\usepackage{wrapfig}
\usepackage{quoting}
\usepackage{graphicx}
\usepackage{calc} 
\usepackage{xspace}

\begin{document}

\newcommand{\NameDataset}[1]{MaCKT\xspace}

\newcommand{\framework}{LLMAgent-CK\xspace}
\newcommand{\singleframework}{LLMAgent-CK-Single\xspace}
\newcommand{\discussframework}{LLMAgent-CK-Discuss\xspace}
\newcommand{\voteframework}{LLMAgent-CK-Vote\xspace}

\newcommand{\metricA}{Question-level Recall\xspace}
\newcommand{\metricB}{Question-level Precision\xspace}
\newcommand{\metricC}{Question-level F1 Score\xspace}

\title{Content Knowledge Identification with Multi-Agent Large Language Models (LLMs)}
\titlerunning{CK Identification with Multi-Agent LLMs}

\author{Kaiqi Yang\inst{1} \and Yucheng Chu\inst{1} \and Taylor Darwin\inst{2} \and Ahreum Han\inst{2} \and Hang Li\inst{1} \and Hongzhi Wen\inst{1} \and Yasemin Copur-Gencturk\inst{2} \and Jiliang Tang\inst{1} \and Hui Liu
\inst{1}}%
\authorrunning{K Yang, et al.}

\institute{Michigan State University 
\email{\{kqyang,chuyuch2,lihang4,wenhongz,tangjili,liuhui7\}@msu.edu}\and
University of Southern California\\
\email{\{tdarwin,ahreumha,copurgen\}@usc.edu}}

\maketitle

\newcommand{\jt}[1]{\textcolor{red}{@JT:~#1@}}
\newcommand{\ycg}[1]{\textcolor{blue}{@Yasemin:~#1@}}
\newcommand{\kq}[1]{\textcolor{purple}{@Korbin:~#1@}}
\newcommand{\yc}[1]{\textcolor{cyan}{@Yucheng:~#1@}}
\newcommand{\hz}[1]{\textcolor{violet}{@Hongzhi:~#1@}}
\newcommand{\hang}[1]{\textcolor{magenta}{@Hang:~#1@}}
\newcommand{\outl}[1]{\textcolor{teal}{\texttt{@Outline:~#1@}\\}}
\newcommand{\del}[1]{}

\begin{abstract}

Teachers' mathematical content knowledge (CK) is of vital importance and need in teacher professional development (PD) programs. Computer-aided asynchronous PD systems are the most recent proposed PD techniques, which aim to help teachers improve their PD equally with fewer concerns about costs and limitations of time or location. However, current automatic CK identification methods, which serve as one of the core techniques of asynchronous PD systems, face challenges such as diversity of user responses, scarcity of high-quality annotated data, and low interpretability of the predictions. To tackle these challenges, we propose a Multi-{Agent} {LLM}s-based framework, \framework, to assess the user responses’ coverage of identified CK learning goals without human annotations. By taking advantage of multi-agent LLMs in strong generalization ability and human-like discussions, our proposed \framework presents promising CK identifying performance on a real-world mathematical CK dataset \texttt{\NameDataset}. Moreover, our case studies further demonstrate the working of the multi-agent framework.


\end{abstract}

\section{Introduction\label{sec:intro}}



Professional Development (PD) plays a crucial role in keeping educators informed about current trends in education and providing them with the necessary knowledge and skills for effective teaching~\cite{copur2016sustainable,desimone2009improving,penuel2007makes}. This is especially vital for subjects such as mathematics, as teachers may enter the profession with different levels of preparation in the subject-specific Content Knowledge (CK)~\cite{heubeck2022emergency,copur2023promising}. Historically, PD for educators is primarily conducted with human experts via in-person or synchronous online meetings. However, this strategy can be costly and not available for \textit{all} teachers, especially those in rural areas~\cite{esquibel2023teacher}, and creates disparities in the access to high-quality PD programs~\cite{powell2019teachers,glover2016investigating}. In addition, the need for timely and efficient evaluation and feedback exacerbates the challenge, especially for the programs lacking real-time PD facilitators. The aforementioned challenges have motivated computer-aided asynchronous PD as an alternative to traditional human-driven PD. By propagating and downloading programs that process user responses automatically, teachers can improve their PD equally with fewer concerns about costs and limitations of time or location\cite{burns2023barriers,fishman2013comparing,unesco2023technology}. 

Within asynchronous PD programs, the most challenging process is to inspect teachers' mastered CK, as it requires both deep understanding with domain-specific knowledge and precise CK identification capabilities to teachers' responses. To achieve this goal, prior systems commonly use free-text questions to test teachers' CK and utilize rule-based systems, traditional machine learning methods or deep learning models to automatically identify teachers' CK by analyzing their responses\footnote{A formal definition for the problem of content knowledge identification is provided in Section\ref{sec:problem}.}\cite{copur2023promising,yoo2023using,kersting2014automated}. However, these CK identifying methods have their own limitations: (1) free text responses are diversified, the static matching rules fail to discern the semantic meaning of the text, which causes low performances; (2) the success of traditional machine learning methods heavily relies on the size and quality of the available training dataset. As the CK identification task is domain-specific, annotating massive high-quality data is costly, which limits the usage of machine learning methods; (3) the emerging deep learning models, especially pre-trained models, despite largely addressing the aforementioned challenges, still suffer from poor interpretability. This impedes their wide usage in education scenarios. 

To tackle these challenges, in this paper, we propose a multi-agent LLMs-based framework, \framework, to identify the user responses’ coverage of the CK learning goals. By taking advantage of the Large Language Model's (LLM) broad knowledge learned during the pre-training phase~\cite{chang2023survey}, \framework has strength in generating strong results without labeled data. This equips the LLM-based model with an overwhelming advantage over the traditional machine learning or deep learning methods, which must be trained on massive labeled data. Apart from that, by introducing discussions between multiple LLMs, \framework eliminates the potential bias of results from a single model and provides a better alignment with the traditional human expertise feedback. At last, since LLMs are generative models, \framework provides both identified results and their associated reasons within the outputs. The generated reasons not only provide more valuable information for the following steps of PD but also give more confidence to the educational users. To demonstrate the effectiveness of \framework, we conduct comprehensive experiments with a real-world mathematical CK dataset \texttt{\NameDataset}. In our experiment, we find \framework can achieve up to 95.83\% precision scores, which is comparable with the human annotators. Our further case studies over one implementation of \framework (namely, \discussframework as introduced in Section~\ref{sec:frame1}), demonstrate the effectiveness of the multi-agent framework.

\section{Related Works}

\subsection{Automatic Short Answer Grading}


Although the study on teacher response identification is rarely explored, Automatic Short Answer Grading (ASAG)\cite{zhang2020going,camus2020investigating} is a well-established research topic focusing on scoring student answers according to reference answers. Early studies \cite{siddiqi2008systematic,jordan2012short} perform word- and phrase-level pattern matching to grade student answers. \cite{wang2008assessing} represents both answer and reference text with TF-IDF~\cite{Joachims1997APA} (term frequency multiplied by inverse document frequency) features, and applies cosine similarity to evaluate the student's responses based on the reference answers. With the rise of deep learning algorithms, advanced text embedding techniques, such as word2vec \cite{mikolov2013efficient} and BERT \cite{devlin2018bert}, have been widely adopted in studies to enhance the semantic representations of answer and reference text \cite{gomaa2020ans2vec}. Additionally, studies by \cite{liu2019automatic,uto2020neural} also explore building advanced neural network designs to catch the deep semantic relationships between the answer and reference text. At last, with the emergence of LLMs, recent works \cite{schneider2023towards,nicula2023automated} have started to adopt LLM with prompt-tuning tricks \cite{brown2020language} for ASAG tasks. 


\subsection{Multi-Agent Large Language Models}

Multi-agent LLMs \cite{wu2023autogen,chan2023chateval} is a most recent proposed LLM enhancing technique, which involves the fusion of multiple LLM-driven agents working cooperatively. Unlike the traditional single model that responds directly, multi-agent systems consist of various AI agents \cite{shinn2023reflexion}, each specializing in distinct domains and contributing to comprehensive problem-solving. This collaborative synergy results in more nuanced and effective solutions. \cite{li2023camel} proposes a cooperative agent framework dubbed as role-playing enabling agents to autonomously cooperate to solve complex tasks. \cite{park2023generative} creates a sandbox environment consisting of virtual entities endowed with a character description and memory system. Every intelligent agent is capable of autonomously interacting with other agents and the environment simulating reliable human behavior. \cite{qian2023communicative} establishes a chat-based software development framework that can complete a software design and produce executable software at a reduced cost compared to recruiting human programmers. Besides the collaborations, the multi-agent debate framework has also been explored to enhance the outputs from single-agent systems. \cite{liang2023encouraging} and \cite{du2023improving} use it for translation and arithmetic problems resulting in better results. \cite{chan2023chateval} applies it to evaluate the quality of generated responses on open-ended questions and traditional natural language generation (NLG) tasks, which achieves higher alignment with human assessment. In this work, we design to leverage the multi-agent framework to simulate the discussion and consensus between human judges and generate expert-like feedback to the teacher's responses.

\section{Problem Statement~\label{sec:problem}}
In this work, our goal is to identify CK learning goals reached by the users' responses to a math question. Specifically, given question text $\mathcal{Q}$ consisting of the main question and hints, teachers are asked to provide CK-related answers with free text $\mathcal{T}$. To identify the learning goals reached by the teacher's responses, the system compares $\mathcal{T}$ with expert-designed learning goal list $\mathcal{E}=[e_1, e_2,\dots,e_M]$, where $e_m$ denotes the $m$-th goal. The output of the identification system will be binary responses $O=[o_1, o_2,\dots,o_M]$, where $o_m \in \{0,1\}$ indicates whether $e_m$ is reached by teacher's response.




\section{Method}

In this section, we introduce our multi-agent LLMs conversation framework \framework which can identify the CK learning goals in teachers’ responses. Although a single LLM is powerful and can be directly adopted for the identification task with prompt tuning algorithms \cite{brown2020language,sun2023principle}, such methods could be sub-optimal as the single model's output may have its bias toward some specific perspective. By introducing multiple LLMs with diversified profiles as agents and allowing them to discuss through the conversation, we expect to receive a broad evaluation view and generate a more accurate judgment. In the following subsections, we first give an overview of the framework. Then, we detail the major designs of the framework. At last, we will present three implementations of the framework which simulate the common discussion formats between human experts.



\subsection{An Overview}

\begin{figure}[htbp]
\centering
\includegraphics[width=0.8\textwidth]{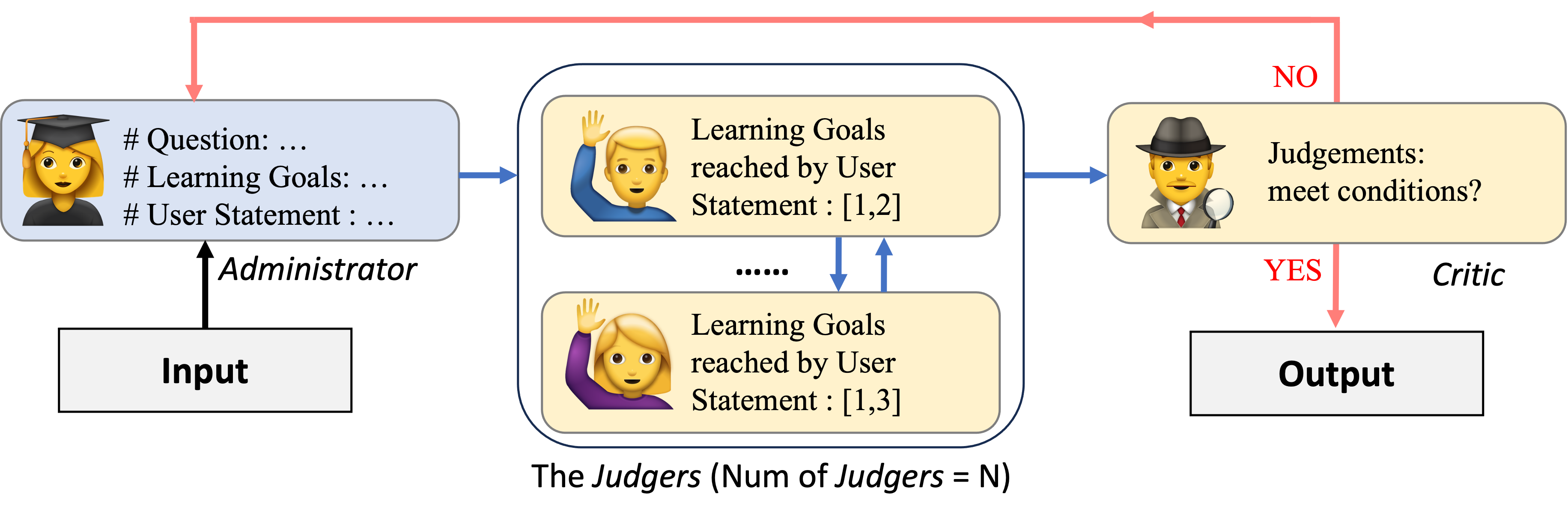} 
\caption{An overview of \framework. It consists of 3 kinds of agent roles: \textit{Administrator}, \textit{Judger}, and \textit{Critic}. The {blue arrows} indicate the \textbf{discussion strategy} for message-passing, controlling how the message is shared between agents. The {red arrows} indicate the \textbf{decision strategy} for decision-making, leading the procedures of the conversation.}
\vspace{-0.2in}
\label{fig:overview}
\end{figure}

An overview of \framework is shown in Figure~\ref{fig:overview}. It consists of three types of LLM-powered agents (i.e. \textit{Administrator}, \textit{Judger}, \textit{Critic}) and two controlling strategies for these agents (i.e. discussion strategy and decision strategy). In general, the \textit{Administrator} agent pre-processes and accepts the input data into the framework, then a group of \textit{Judger} agents (Num. of \textit{Judgers} = $N$, which can be decided by usage goals) makes their answers. After that, the \textit{Critic} agent makes up the final outputs based on the answers from \textit{Judger} agents, and proposes new rounds of conversations if needed. For the controlling strategies, discussion strategies can control the ways of message sharing between agents, while decision strategies guide the \textit{Critic} agent to make final decisions and manage the new conversation. Details of role and controlling strategy design will be discussed in Section~\ref{sec:role} and ~\ref{sec:strategy}, respectively. By adopting different agents and strategies, \framework can simulate many human-being cooperation styles. Therefore, three practical implementations of \framework will be discussed in Section~\ref{sec:imp}. 


\subsection{Role Design~\label{sec:role}}

\textbf{Administrator Agent.} The \textit{Administrator} agent $A_A$ is built on a large language model that can take, preprocess and manage the information from input data, and then broadcast it to other agents when needed.  The input learning goals $\mathcal E$ are the key to the identification task. However, they are provided by human experts, which could be sub-optimal for LLMs to understand given the possible gap between humans and LLMs. Therefore, we assign the ability of rephrasing~\cite{deng2023rephrase} to the \textit{Administrator} agent $A_A$. It always rephrases the input learning goals $\mathcal{E}$, manages other input data, and then broadcasts all the input data (the question $\mathcal Q$, rephrased learning goals $\mathcal E$, and the user's responses $\mathcal T$) to the other agents.

\noindent \textbf{Judger Agent.} The \textit{Judger} agent $A_J$ is designed to identify the learning goals reached by user responses. After receiving the question and list of learning goals from the \textit{Administrator} agent, the \textit{Judger} agent understands and carefully checks the main points of each goal. Then it examines the user responses and identifies which goal is correctly reached. Finally, it outputs the set of learning goals it predicts to be reached by the user responses, as well as explanatory evidence to support the prediction. Therefore, the input prompt for $A_J$ has three sections: 
\begin{itemize}
    \item An introduction to structures of input data ("\textit{Input Data: \# Quesion, [question text]; \# Learning Goals, [goal text]; \# User Response, [user response]...}");
    \item The working flow to solve the problem as mentioned above (e.g. "\textit{Work Flow: 1. split the learning goals; 2. compare the goals; 3. check each goal...}"),
    \item The template of output formats ("\textit{Output Example: \{"evidence": [evidence text], "output":[1,3]\}}")
\end{itemize}

Note that it is optional to broadcast the replies of one $A_J$ to other $A_J$s dependent on the discussion strategy (which will be detailed in Section~\ref{sec:strategy}). If the broadcast is allowed, which simulates the open discussion scenarios in practice, the input prompt will be appended with one more section with the outputs from other \textit{Judger} agents (e.g. "\textit{Judger A: \{"evidence": [evidence text], "output":[1,3]\}; Judger B: \{"evidence": [evidence text], "output":[2,3]...\}}").

\noindent \textbf{Critic Agent.} The \textit{Critic} $A_C$ plays a crucial role in the framework as it collects all outputs from \textit{Judger} agents and checks if there is a consensus reached between these outputs. Given the input $\mathcal Q$ and $\mathcal E$ as references, $A_C$ is asked to check if the outputs from all $A_J$ agents reach an agreement under the guidance of decision strategies (as discussed by \textbf{Decision Strategy} below). If they agree, $A_C$ will terminate the conversation; otherwise, $A_C$ will summarize the reason why disagreement occurs and broadcast it to all the $A_J$ agents, which helps the next round of discussions. New rounds of conversations are called until the $A_C$ terminates it. The final outputs are summarized from individual outputs given by \textit{Judger} agents. 

\subsection{Controlling Strategy Design~\label{sec:strategy}}

\textbf{Discussion Strategy.} The discussion strategy determines whether and how the intermediate outputs of agents can be shared with other agents. For example, we define a strategy named "Open\_Judge", following which the \textit{Judger} agents $A_J$ are aware of the existing outputs from other \textit{Judger} agents when making their own decisions. It simulates the cooperation scenarios where there are open discussions between people, and they are asked to make decisions based on their own reasoning with references to others' opinions. Conversely, a strategy of message-passing named "Closed\_Judge" turns off the answers sharing between \textit{Judger} agents, merely allowing them to communicate with the \textit{Administrator} and \textit{Critic} agent. This strategy can imitate scenarios where people are asked to make decisions independently. 

\noindent\textbf{Decision Strategy.} Now we introduce the strategies of decision-making. The advantage of \framework lies in the group thinking of diverse agents. However, it requires strategies to ensemble the individual outputs and present the consensus as the final outputs. These strategies are often defined for the \textit{Critic}, who serves as the manager of agents. In this work, we design two strategies of decision-making as examples for illustration. The first one is "Total\_Agreement": when all the individual outputs from \textit{Judger} agents are the same, the managing agent \textit{Critic} feels satisfied and terminates the conversation. Otherwise, the \textit{Critic} summarizes the differences and broadcasts them to other agents in preparation for the next round of conversations. The second strategy of decision-making is "Majority\_Voting". When there are enough \textit{Judger} agents, the \textit{Critic} can make final decisions based on the majority consensus of \textit{Judger} agents. Only when there is significant controversy (e.g., the half-half situations) will the \textit{Critic} summarize the decisions and propose new conversations. In short, strategies of decision-making simulate the discussion process when encountering disagreements, with which \textit{Judgers} are asked to discuss more for a consensus.
\subsection{Practical Implementations~\label{sec:imp}}

We demonstrate three practical implementations deriving from the \framework framework. Note that with different numbers of agents and controlling strategies, we can design new implementations of different conversation procedures flexibly, thus simulating different cooperation styles of humans in practice.

\subsubsection{\singleframework~\label{sec:frame0}}
\begin{figure}[htbp]
\centering
\includegraphics[width=0.9\textwidth]{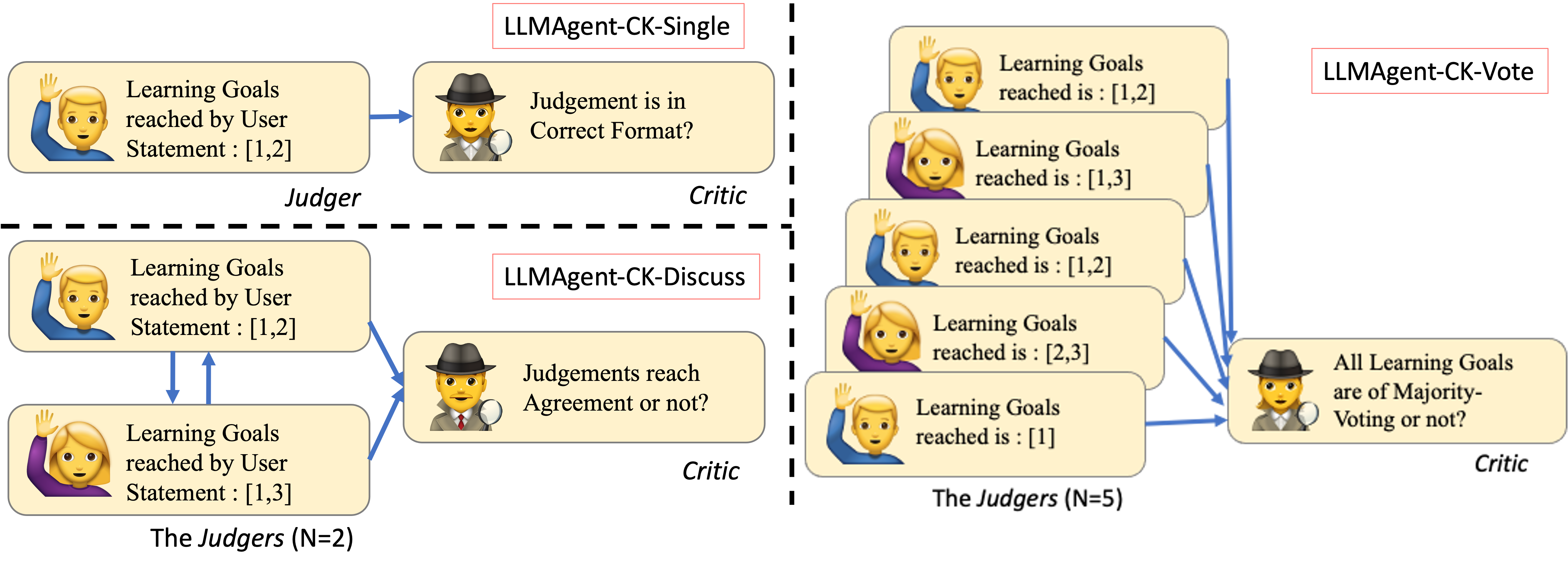} 
\caption{An illustration of three implementations of \framework: \singleframework (left-top), \discussframework(left-bottom) and \voteframework(right). In the figure, the information is passed by blue arrows (rule of message-passing).}
\vspace{-0.2in}
\label{fig:imp}
\end{figure}

We first show a naive implementation of \framework, namely, \singleframework as shown in the left-top corner of Figure~\ref{fig:imp}. This implementation simulates the scenario where people make decisions individually, so the number of \textit{Judger} agents $A_J$ is set as $N=1$. As there is \textbf{only one \textit{Judger}}, the conversation between \textit{Judger} agents is unnecessary and the critic process degenerates to only check if the outputs are in the correct format. The goal of this implementation is to check the ability of the single LLM to solve CK identification problems, which serves as a baseline model for other LLM-based implementations. 

\subsubsection{\discussframework~\label{sec:frame1}}

Our second implementation is \discussframework, which includes \textbf{two \textit{Judger} agents} $A_J$ and allows them to communicate with each other when making decisions. This implementation simulates a public-discussion scenario, where all the participants are aware of others' opinions, and are asked to make decisions based on both their own rationale and others' opinions. The procedure of \discussframework is shown in the left-bottom corner of Figure~\ref{fig:imp}. To be specific, the \textbf{ controlling strategies} are set as the following. \uline{Open\_Judge}: when the \textit{Judger} $A_J$ is making up answers, it first receives the answers from other \textit{Judger} agents (if any) in addition to the input data. \uline{Total\_Agreement}: when \textit{Critic} $A_C$ finds any disagreement between answers of \textit{Judger} agents, it summarizes the disagreement and informs the \textit{Judger} agents $A_J$, starting the next round of discussions.

\subsubsection{\voteframework~\label{sec:frame2}}
In our third implementation of \voteframework, we incorporate \textbf{five \textit{Judger} agents} $A_J$, with a \textit{Critic} $A_C$ who decides on agreements through majority voting. The procedure of \voteframework is shown on the right side of Figure~\ref{fig:imp}. Different from \discussframework, \voteframework simulates real-world majority voting activities based on independent individual decisions. We set \textbf{controlling strategies} as follows. \uline{Closed\_Judge}: all $A_J$ agents make up answers independently, without knowing answers from other $A_J$ agents. \uline{Majority\_Voting}: for the Critic $A_C$, it decides the final outputs by majority-voting mechanism. With 5 answers from \textit{Judger} agents, if a learning goal $e$ is mentioned (ignored) by at least 4 \textit{Judger} agents, \textit{Critic} $A_C$ will add (remove) it in the final decision. Otherwise, \textit{Critic} $A_C$ summarizes the contention, informs the \textit{Judger} agents, and proposes a new round of discussions. 

\section{Experiment}
In this section, we illustrate the experiments of the proposed LLM-based multi-agent conversation framework. First, we introduce the experimental settings. Then, we discuss the results of the three implementations on the math CK dataset. Finally, we perform a case study to illustrate how multiple agents perform collaborative work.

\subsection{Experimental Seting~\label{sec:setup}}
\subsubsection{Dataset~\label{sec:Dataset}} 
The data used in this study came from a project that was designed to investigate the impact of a computerized professional development (PD) program on teachers and their students. The project collected the data of open-text responses to content knowledge problems that the participating teachers provided when they were interacting with the system. The teachers whose data were used in this study were middle school mathematics teachers. The majority of these teachers were female (81\%) and White (69\%), similar to the demographic profile of the U.S. teaching workforce. To ensure high-quality data, we construct a dataset \texttt{\NameDataset}~based on a set of usage data of the above asynchronous PD program with the selected topics of math CK.

As stated in Section~\ref{sec:problem}, the analytic data contains questions, learning goals, and user responses, while the expected output is the lists of reached learning goals. To get the ground-truth labels of the reached goals, we recruit domain experts to discuss and annotate the dataset. Below is one example:
\begin{itemize}
    \item \textbf{Questions} "\textit{All the workers paint at the same rate, and 6 workers can paint one house in 7 hours. Is the relationship between the number of workers and time to paint the house proportional or not proportional?}". 
    \item \textbf{Learning Goals}: "\textit{(1) as the number of workers changes, the time it takes to paint a house also changes}", "\textit{(2) the amount of time it takes to paint the houses for the number of workers is not constant...}", "\textit{(3) ... the relationship is not proportional}".
    \item \textbf{User Response}: "\textit{...for 6 workers it would take 7 hours, for 12 workers it would take 3.5 hours.}".
    \item \textbf{Expert Annotation}: \textit{[1,2]}.
\end{itemize} 

 From the \texttt{\NameDataset}~dataset, we select a total of 9 sample questions to conduct experiments, including 452 user responses. To simplify the expression, we use "$Q1, Q2, ..., Q9$" to denote these math questions. The statistics of questions are shown in Table~\ref{tab:questions}. Note that the ground-truth labels are only utilized for the evaluation purpose. In other words, \framework does not leverage these labels in the predictions.
 
\begin{table}[!htbp]
\caption{Statistics of \texttt{\NameDataset{}} Dataset. The Question Abbr. are order numbers of questions.}
\label{tab:questions}
\centering
\begin{tabular}{c|ccccccccccc}
\midrule
\textbf{Question Abbr.} & \textbf{Q1} & \textbf{Q2} & \textbf{Q3} & \textbf{Q4} & \textbf{Q5} & \textbf{Q6} & \textbf{Q7} & \textbf{Q8} & \textbf{Q9} & \textbf{Total}\\\hline
\textbf{\# User Responses} & 60 & 69 & 50 & 36 & 55 & 22 & 37 & 48 & 75 & \uline{452} \\
\midrule
\end{tabular}
\vspace{-15pt}
\end{table}

\subsubsection{Metrics}
Recall that the outputs of problems are the lists of binary responses $O=[o_1, o_2,\dots,o_M]$, where $o_m \in \{0,1\}$ indicates whether learning goal $e_m$ is reached or not. We formally define three metrics as below. Given a question having $J$ user responses $\mathcal{T}$; for an example of user response $\mathcal{T}_j$, assuming the true output is $O_j$ and the prediction is $\hat O_j$: 
\begin{align*}
    &\text{\metricA}=\frac{\sum_j^J|O_j\cap\hat O_j|}{\sum_j^J|O_j|};\\&\text{\metricB}=\frac{\sum_j^J|O_j\cap\hat O_j|}{\sum_j^J|\hat O_j|};\\
    &\text{\metricC}=2*\frac{\text{\metricA}*\text{\metricB}}{\text{\metricA}+\text{\metricB}};
\end{align*}
where $|O_j\cap\hat O_j|$ is the number of goals correctly identified by models. The \metricA represents the proportion of correctly identified goals out of the total number of goals that are \textit{expected} to be identified. This metric quantifies the extent to which the goals annotated by experts are successfully identified by the prediction. \metricB calculates the ratio of correctly identified goals to \textit{all} identified goals, thereby indicating the ratio of the model's predictions in alignment with expert annotations to all predicted goals.

\subsubsection{Baselines}

Note that our framework follows the zero-shot setting\cite{kojima2205large} and ground truth labels are not used for training \framework. Thus, the majority of existing machine learning or deep learning methods are not applicable in this setting. We define two semantic matching baselines. The process is: we use pre-trained language models to obtain semantic embeddings for two sections of text: one is the concatenation of the question and learning goal, and the other is the user response. Then we calculate their cosine similarities. We claim a match if the similarity is larger than a predefined threshold; otherwise, we regard the goal as unfulfilled by user responses. In this work, the two pre-trained models we choose to generate embeddings are BERT~\cite{devlin2018bert} and Sentence BERT (S\_BERT)~\cite{reimers2019sentence}. 

\subsubsection{Implementation Details}

For \framework, we choose the flagship models of the GPT family, i.e. the \texttt{GPT-3.5 Turbo} and \texttt{GPT-4 Turbo}, as the backbone LLMs. To set the agent roles or "persona" diverse~\cite{white2023prompt}, we select the temperature parameters for LLMs from $0$ to $0.7$. For the prompts of agents, we follow the zero-shot (no labeled data) setting, where no other samples from the dataset are shown to the models. Lastly, in our setting of three implementations, no annotations are provided to the models.
\subsection{Results}

\subsubsection{Single-Agent Baseline Comparison~\label{sec:single}}
We first show the performance of the single agent settings, i.e. the baselines and \singleframework. Although they follow different frameworks, all of them rely on one model (agent) to identify learning goals, without the conversation mechanism between multiple agents. As shown in Table~\ref{tab:baseline}, both two state-of-the-art GPT models achieve superior performances in identifying math CK learning goals for all 9 questions, especially when considering the agreement across the education experts on most of the questions is around 85\% to 90\%. One notable phenomenon is that the advanced \texttt{GPT-4} is not always more powerful than \texttt{GPT-3.5} on the \metricA metric. However, the \metricB of \texttt{GPT-4} are much more consistent and better than that of \texttt{GPT-3.5}. In addition, \texttt{GPT-4} achieves better on the \metricC on 8 out of 9 questions.

As for semantic matching baselines, although we empirically selected the threshold with the best F1 scores, their performances are consistently poorer than GPT models. Notably, even though the \metricA of baselines appears high, there is a significant discrepancy between high recall scores and notably lower precision scores. This suggests: (1) the baselines tend to indiscriminately classify learning goals as being reached, suggesting a limited capacity due to the complexity of the input data; and (2) the majority of learning goals are indeed marked as unreached by the user responses, highlighting an imbalance in the distribution of reached and unreached goals that further complicates the task.

\begin{table}[!htbp]
\vspace{-5pt}
\caption{Performance of the baselines and \singleframework on identifying CK. In this table, GPT v3.5 and GPT v4 stand for the \singleframework with the corresponding version of GPT; the BERT and S\_BERT indicate two baselines as introduced in Section~\ref{sec:setup}. For each metric and each question, the best performance is marked with \textbf{bold font}.}
\label{tab:baseline}
\centering
\resizebox{\textwidth}{!}{
\begin{tabular}{c|c|ccccccccc|c}  
\hline
\textbf{Metric} & \textbf{Model} & \textbf{Q1} & \textbf{Q2} & \textbf{Q3} & \textbf{Q4} & \textbf{Q5} & \textbf{Q6} & \textbf{Q7} & \textbf{Q8} & \textbf{Q9} & \textbf{Ave} \\

\cline{1-12} 

\multirow{4}{*}{\begin{tabular}[c]{@{}l@{}}Question-level\\ Recall\end{tabular}} & BERT & \textbf{95.06} & \textbf{100} & \textbf{100} & \textbf{95.65} & \textbf{94.37} & \textbf{100} & \textbf{92.31} & \textbf{97.22} & \textbf{98.15} & \textbf{96.97} \\
 & S\_BERT & 87.65 & 87.50 & 88.31 & 97.83 & 87.32 & 95.65 & 100 & 87.50 & 90.74 & 91.39 \\
 & GPT v3.5 & 75.31 & 87.01 & 78.57 & 80.43 & 84.51 & 39.13 & 67.31 & 63.89 & 75.93 & 72.45 \\
 & GPT v4 & 64.20 & 87.01 & 80.36 & 54.35 & 64.79 & 86.96 & 65.38 & 61.11 & 62.96 & 69.68 \\
 
 \cline{1-12} 
 
\multirow{4}{*}{\begin{tabular}[c]{@{}l@{}}Question-level\\ Precision\end{tabular}} & BERT & 47.53 & 29.63 & 53.47 & 45.83 & 44.97 & 52.27 & 48.98 & 52.63 & 24.65 & 44.44 \\
 & S\_BERT & 48.30 & 32.67 & 55.28 & 46.88 & 46.97 & 52.38 & 48.15 & 52.07 & 25.79 & 45.39 \\
 & GPT v3.5 & 57.55 & 70.53 & 41.12 & 63.79 & 61.86 & 40.91 & 67.31 & 69.70 & 41.00 & 57.09 \\
 & GPT v4 & \textbf{81.25} & \textbf{80.72} & \textbf{86.53} & \textbf{75.75} & \textbf{95.83} & \textbf{90.9} & \textbf{89.47} & \textbf{89.79} & \textbf{62.96} & \textbf{83.69} \\
 
 \cline{1-12} 
 
\multirow{4}{*}{\begin{tabular}[c]{@{}l@{}}Question-level\\ F1\end{tabular}} & BERT & 63.37 & 45.71 & 69.68 & 61.97 & 60.91 & 68.66 & 64.00 & 68.29 & 39.41 & 60.22 \\
 & S\_BERT & 62.28 & 47.57 & 68.00 & \textbf{63.38} & 61.08 & 67.69 & 65.00 & 65.28 & 40.16 & 60.05 \\
 & GPT v3.5 & 65.24 & 77.91 & 53.99 & 71.15 & 71.43 & 40.00 & 67.31 & 66.67 & 53.25 & 62.99 \\
 & GPT v4 & \textbf{71.73} & \textbf{83.75} & \textbf{83.33} & 63.29 & \textbf{77.31} & \textbf{88.89} & \textbf{75.55} & \textbf{72.72} & \textbf{62.96} & \textbf{75.50}\\

\hline

\end{tabular}}
\vspace{-30pt}
\end{table}

\subsubsection{Multi-Agent Comparison~\label{sec:mulresult}}
Now we compare the performances of the multi-agent implementations. Based on our prior experiment, we select \texttt{GPT-4 Turbo} as the LLM backbone. In addition, without loss of generality, we select 4 of the 9 questions from the dataset, i.e. $Q1$, $Q4$, $Q7$, and $Q8$. Note that each selected question has one of the lowest \metricA scores with \texttt{GPT-4 Turbo} under \singleframework. We made such a selection to check the effects of multi-agent frameworks on the hard problems for LLMs. With the comparison in this subsection, we find that the multi-agent frameworks help to further improve the performance of LLMs, even when the individual model works poorly. Besides, for the CK identification problem, we find that the voting mechanism leads to even better results than the discussion.

\begin{figure}[!thbp]
\vspace{-0.15in}
    \centering
    \includegraphics[width=0.9\textwidth]{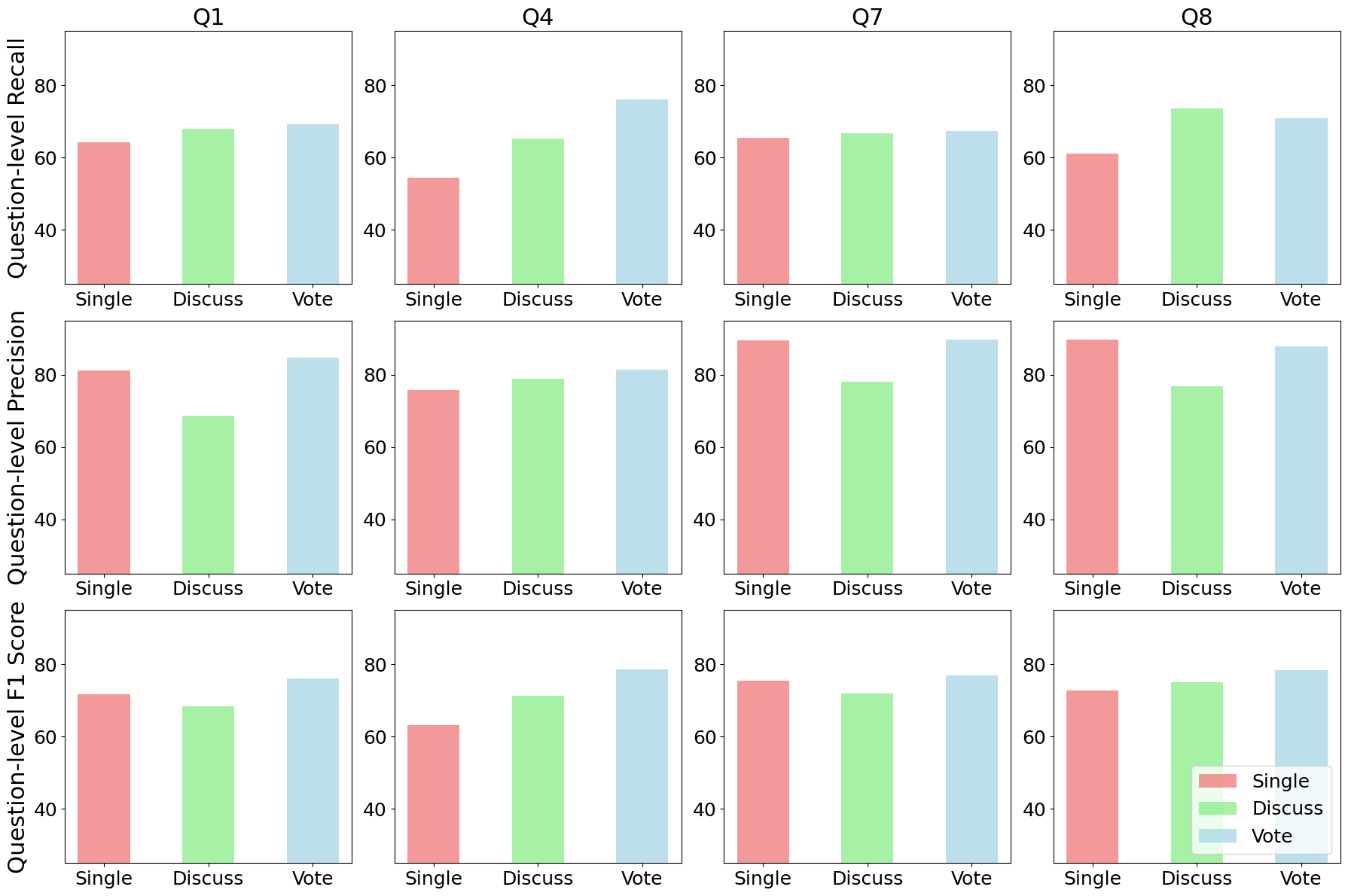}
\vspace{-0.1in}
    \caption{Performance of \discussframework (green) and \voteframework (blue). We also show the results of \singleframework (red) as the reference.}
    \vspace{-0.4in}
    \label{fig:conversation_figure}
\end{figure}

The performance of \discussframework is shown in Figure~\ref{fig:conversation_figure}. As observed, when using the multi-agent discussion, most of the performances improved compared with the single-agent setting. For example, for the question $Q4$, the three metrics increase from $54.35\%$, $75.75\%$, $63.29\%$ (\singleframework with \texttt{GPT-4 Turbo}) to $65.22\%$, $78.95\%$, and $71.43\%$. Although in a few settings, such as \metricC of $Q1$ and $Q7$, the performances marginally drop, in most cases the multi-agent frameworks help to remedy the limitation of the single LLM. 

In Figure~\ref{fig:conversation_figure}, we show the results of \voteframework, the implementation with majority-voting mechanisms. As observed, the \voteframework achieves higher scores on almost all four selected questions. Especially, the \metricC of four questions improves by +7.87\%, +7.23\%, +5.01\%, +3.28\% compared with \discussframework. As \metricC indicates a more balanced view of the performance, this indicates that the \voteframework design is more powerful than \discussframework.




\vspace{-0.05in}
\section{Case Study}
\vspace{-0.1in}

In this section, we illustrate one case study of the discussion process to demonstrate the advantage of \discussframework. This question asks which one of the four rectangles (7 by 10, 17 by 20, 27 by 30, 37 by 40) looks the most like a square, and the example user response is "\textit{I believe the rectangle that looks most square is the rectangle that is 37 feet by 40 feet, because it is larger than the rest of the rectangles.}". To save space, we only show parts of the conversation between the \textit{Judger} agents and the \textit{Critic} in Table~\ref{tab:casestudy}. 

In the first round of conversation, \textit{Judger A} wrongly thought learning goal 3 ("\textit{the answer is 37 feet by 40 feet}") was covered by the user response, as it was misled by the user response which accidentally selected the correct answer ("\textit{...is the rectangle that is 37 feet by 40 feet}") but supported by wrong reasons (i.e. it focused on the size rather than the ratio of sides).  Conversely, \textit{Judger B} accurately discerned the situation and rendered an appropriate judgment. Through the exchange of rationale and collective summarization, both \textit{Judger A} and \textit{Judger B} converged on correct conclusions. Consequently, despite only one of the two \textit{Judger} agents making correct judgments at first, the collaborative conversation effectively ensures the build-up of correct output with group cooperation.

\begin{table}[!thbp]
\vspace{-0.2in}
\caption{A case study of conversations of \discussframework. In this table, "\textit{[input data]}" indicates the input data introduced in Section~\ref{sec:Dataset}; "\textit{[two rows above]}" indicates the content in two rows above, as \textit{Critic} is collecting answers from \textit{Judgers}.}
\label{tab:casestudy}
\centering
\begin{tabular}{p{0.12\textwidth}|p{0.85\textwidth}}
\midrule
\textbf{Agent} & \textbf{Message} \\
\hline
Admin & \textit{[input data]} \\
\hline
Judger A & \{"evidence": "The user mentioned the size of the rectangle (37 feet by 40 feet) as the reason it looks most like a square.", "output": [3]\} \\
\hline
Judger B & \{"evidence": "The user stated that the 37 feet by 40 feet rectangle looks most like a square due to its size but did not provide reasoning related to the ratio of width to length or the sides difference.", "output": [ ]\} \\
\hline
Critic & The 'output' lists are not the same. The first 'output' is [3], while the second 'output' is an empty list. The difference is that one judge assistant identified evidence related to the learning goal, while the other did not find the user's reasoning aligned with the expected evidence. CONTINUE. \\
\hline
Admin & \textit{[input data]} \\
\hline
Judger A & \{"evidence": "The user incorrectly used the overall size ('larger than the rest of the rectangles') as a reason for the rectangle looking more like a square, which does not align with the expected reasoning of width to length ratio or side difference.", "output": [ ]\} \\
\hline
Judger B & \{"evidence": "The user stated that the 37 feet by 40 feet rectangle looks most like a square due to its size but did not provide reasoning related to the ratio of width to length or the sides difference.", "output": [ ]\} \\
\hline
Critic & The 'output' lists are the same, both are empty lists. The shared 'output' is "[ ]". The replies from the judge assistants are: \textit{[two rows above]}. AGREEMENT. \\
\midrule
\end{tabular}
\end{table}

\vspace{-0.3in}
\section{Conclusion}
\vspace{-0.1in}

In this paper, we propose a Multi-Agent LLMs-based framework, \framework, to solve the challenging CK identification problem in computer-aided asynchronous PD systems. With comprehensive experiments over the real-world mathematical CK dataset, \texttt{\NameDataset}, we demonstrate its capabilities in generating promising identification results without the need for annotations. Furthermore, the case study showcasing the human-like correction capabilities of \framework underscores its considerable potential to achieve results that closely align with human judgment. In our future research, we aim to explore a broader range of communication strategies to improve its overall performance and efficiency. Moreover, as the field of PD continues to advance and yield more data, we plan to employ advanced fine-tuning techniques to elevate the framework's effectiveness.

\bibliographystyle{splncs04}
\bibliography{ref}

\begin{thebibliography}{10}
\providecommand{\url}[1]{\texttt{#1}}
\providecommand{\urlprefix}{URL }
\providecommand{\doi}[1]{https://doi.org/#1}

\bibitem{brown2020language}
Brown, T., et~al.: Language models are few-shot learners. NeurIPS  (2020)

\bibitem{burns2023barriers}
Burns, M., et~al.: Barriers and supports for technology integration: Views from teachers. UNESCO Global Monitoring Report.  (2023)

\bibitem{camus2020investigating}
Camus, L., Filighera, A.: Investigating transformers for automatic short answer grading. In: AIED 2020. Springer (2020)

\bibitem{chan2023chateval}
Chan, C.M., et~al.: Chateval: Towards better llm-based evaluators through multi-agent debate. arXiv preprint arXiv:2308.07201  (2023)

\bibitem{chang2023survey}
Chang, Y., et~al.: A survey on evaluation of large language models. ACM TIST  (2023)

\bibitem{copur2023promising}
Copur-Gencturk, Y., Orrill, C.H.: A promising approach to scaling up professional development: intelligent, interactive, virtual professional development with just-in-time feedback. J. Math. Teach. Educ.  (2023)

\bibitem{copur2016sustainable}
Copur-Gencturk, Y., Papakonstantinou, A.: Sustainable changes in teacher practices: A longitudinal analysis of the classroom practices of high school mathematics teachers. J. Math. Teach. Educ.  (2016)

\bibitem{deng2023rephrase}
Deng, Y., et~al.: Rephrase and respond: Let large language models ask better questions for themselves. arXiv preprint arXiv:2311.04205  (2023)

\bibitem{desimone2009improving}
Desimone, L.M.: Improving impact studies of teachers’ professional development: Toward better conceptualizations and measures. Educational researcher  (2009)

\bibitem{devlin2018bert}
Devlin, J., et~al.: Bert: Pre-training of deep bidirectional transformers for language understanding. arXiv preprint arXiv:1810.04805  (2018)

\bibitem{du2023improving}
Du, Y., et~al.: Improving factuality and reasoning in language models through multiagent debate. arXiv preprint arXiv:2305.14325  (2023)

\bibitem{esquibel2023teacher}
Esquibel, J.S., Darwin, T.: The teacher talent pipelines: A systematic literature review of rural teacher education in the virtual age. Handbook of research on advancing teaching and teacher education in the context of a virtual age  (2023)

\bibitem{fishman2013comparing}
Fishman, B., et~al.: Comparing the impact of online and face-to-face professional development in the context of curriculum implementation. J. Teach. Educ.  (2013)

\bibitem{glover2016investigating}
Glover, T.A., et~al.: Investigating rural teachers' professional development, instructional knowledge, and classroom practice. J. Res. Rural Educ.cation  (2016)

\bibitem{gomaa2020ans2vec}
Gomaa, W.H., Fahmy, A.A.: Ans2vec: A scoring system for short answers. In: AMLTA 2019. Springer (2020)

\bibitem{heubeck2022emergency}
Heubeck, E.: Emergency certified teachers: Are they a viable solution to shortages? Education Week  (2022)

\bibitem{Joachims1997APA}
Joachims, T.: A probabilistic analysis of the rocchio algorithm with tfidf for text categorization. In: ICML (1997)

\bibitem{jordan2012short}
Jordan, S.: Short-answer e-assessment questions: Five years on. Int. Comput. Assist. Assess. Conf.  (2012)

\bibitem{kersting2014automated}
Kersting, N.B., Sothers: Automated scoring of teachers’ open-ended responses to video prompts: Bringing the classroom-video-analysis assessment to scale. Educ. Psychol. Meas.  (2014)

\bibitem{kojima2205large}
Kojima, T., Gu, S.S., Reid, M., Matsuo, Y., Iwasawa, Y.: Large language models are zero-shot reasoners, 2022. URL https://arxiv. org/abs/2205.11916

\bibitem{li2023camel}
Li, G., et~al.: Camel: Communicative agents for" mind" exploration of large scale language model society. arXiv preprint arXiv:2303.17760  (2023)

\bibitem{liang2023encouraging}
Liang, T., et~al.: Encouraging divergent thinking in large language models through multi-agent debate. arXiv preprint arXiv:2305.19118  (2023)

\bibitem{liu2019automatic}
Liu, T., et~al.: Automatic short answer grading via multiway attention networks. In: AIED 2019. Springer (2019)

\bibitem{mikolov2013efficient}
Mikolov, T., et~al.: Efficient estimation of word representations in vector space. arXiv preprint arXiv:1301.3781  (2013)

\bibitem{nicula2023automated}
Nicula, B., et~al.: Automated assessment of comprehension strategies from self-explanations using llms. Information  (2023)

\bibitem{park2023generative}
Park, J.S., et~al.: Generative agents: Interactive simulacra of human behavior. In: ACM UIST (2023)

\bibitem{penuel2007makes}
Penuel, W.R., et~al.: What makes professional development effective? strategies that foster curriculum implementation. Am. Educ. Res. J.  (2007)

\bibitem{powell2019teachers}
Powell, C.G., Bodur, Y.: Teachers’ perceptions of an online professional development experience: Implications for a design and implementation framework. Teaching and Teacher Education  (2019)

\bibitem{qian2023communicative}
Qian, C., et~al.: Communicative agents for software development. arXiv preprint arXiv:2307.07924  (2023)

\bibitem{reimers2019sentence}
Reimers, N., Gurevych, I.: Sentence-bert: Sentence embeddings using siamese bert-networks. arXiv preprint arXiv:1908.10084  (2019)

\bibitem{schneider2023towards}
Schneider, J., et~al.: Towards llm-based autograding for short textual answers. arXiv preprint arXiv:2309.11508  (2023)

\bibitem{shinn2023reflexion}
Shinn, N., Labash, B., Gopinath, A.: Reflexion: an autonomous agent with dynamic memory and self-reflection. arXiv preprint arXiv:2303.11366  (2023)

\bibitem{siddiqi2008systematic}
Siddiqi, R., Harrison, C.: A systematic approach to the automated marking of short-answer questions. In: IEEE INMIC. IEEE (2008)

\bibitem{sun2023principle}
Sun, Z., et~al.: Principle-driven self-alignment of language models from scratch with minimal human supervision. arXiv preprint arXiv:2305.03047  (2023)

\bibitem{unesco2023technology}
UNESCO: Global education monitoring report 2023: Technology in education -- a tool on whose terms? Tech. rep., UNESCO (2023)

\bibitem{uto2020neural}
Uto, M., Xie, Y., Ueno, M.: Neural automated essay scoring incorporating handcrafted features. In: COLING (2020)

\bibitem{wang2008assessing}
Wang, H.C., Chang, C.Y., Li, T.Y.: Assessing creative problem-solving with automated text grading. Comput. Educ.  \textbf{51}(4),  1450--1466 (2008)

\bibitem{white2023prompt}
White, J., et~al.: A prompt pattern catalog to enhance prompt engineering with chatgpt. arXiv preprint arXiv:2302.11382  (2023)

\bibitem{wu2023autogen}
Wu, Q., et~al.: Autogen: Enabling next-gen llm applications via multi-agent conversation framework. arXiv preprint arXiv:2308.08155  (2023)

\bibitem{yoo2023using}
Yoo, J., Kim, M.K.: Using natural language processing to analyze elementary teachers’ mathematical pedagogical content knowledge in online community of practice. Contemporary Educational Technology  (2023)

\bibitem{zhang2020going}
Zhang, Y., Lin, C., Chi, M.: Going deeper: Automatic short-answer grading by combining student and question models. UMUAI  \textbf{30} (2020)

\end{thebibliography}

\end{document}